\documentclass[conference]{IEEEtran} 
\ifCLASSOPTIONcompsoc
  \usepackage[nocompress]{cite}
\else
  \usepackage{cite}
\fi
\ifCLASSINFOpdf
\else
\fi
\usepackage[colorlinks=true,
            linkcolor=red,
            urlcolor=blue,
            citecolor=blue]{hyperref}
\usepackage{amsmath}
\usepackage{graphicx}
\usepackage{mathtools}
\usepackage{relsize}
\usepackage{graphicx}
\usepackage{algorithm}
\usepackage{filecontents,lipsum}
\usepackage{cite}
\usepackage{comment}
\usepackage{multirow}
\usepackage{amsfonts}
\usepackage[noend]{algpseudocode}
\usepackage[utf8]{inputenc}
\usepackage{graphicx}
\usepackage{caption}
\usepackage{latexsym}
\usepackage{etoolbox}
\usepackage{subcaption, float}
\makeatletter
\usepackage[inline]{enumitem}
\def\BState{\State\hskip-\ALG@thistlm}
\makeatother

\AfterEndEnvironment{table}{\vskip-1ex}

\begin{document}
%
\title{Analysis of Railway Accidents' Narratives Using Deep Learning}
%
%
%

\author{Mojtaba Heidarysafa\IEEEauthorrefmark{1}, 
\IEEEauthorblockN{Kamran Kowsari\IEEEauthorrefmark{1}, Laura E. Barnes\IEEEauthorrefmark{1}\IEEEauthorrefmark{3}, and Donald E. Brown\IEEEauthorrefmark{1}\IEEEauthorrefmark{3}}

\IEEEauthorblockA{\IEEEauthorrefmark{1} Department of System and Information Engineering,
University of Virginia,
Charlottesville, VA, USA}

\IEEEauthorblockA{\IEEEauthorrefmark{3} Data Science Institute, 
University of Virginia,
Charlottesville, VA, USA}

\{\href{mailto:mh4pk@virginia.edu}{mh4pk},
\href{mailto:kk7nc@virginia.edu}{kk7nc},
\href{mailto:lb3dp@virginia.edu}{lb3dp},
\href{mailto:deb@virginia.edu}{deb}\}@virginia.edu}

%
%

\markboth{IEEE TRANSACTIONS ON INTELLIGENT TRANSPORTATION SYSTEMS, VOL. XX, NO. XX, FEBRUARY 2018}%
{Shell \MakeLowercase{\textit{et al.}}: Bare Demo of IEEEtran.cls for Journals}
%



\maketitle

\begin{abstract}
Automatic understanding of domain specific texts in order to extract useful relationships for later use is a non-trivial task. One such relationship would be between railroad accidents' causes and their correspondent descriptions in reports. From~2001 to 2016 rail accidents in the U.S. cost more than~\$4.6B. Railroads involved in accidents are required to submit an accident report to the Federal  Railroad  Administration~(FRA).
These reports contain a variety of fixed field entries including primary cause of the accidents (a coded variable with~389 values) as well as a narrative field which is a short text description of the accident.
Although these narratives provide more information than a fixed field entry, the terminologies used in these reports are not easy to understand by a non-expert reader. Therefore, providing an assisting method to fill in the primary cause from such domain specific texts~(narratives) would help to label the accidents with more accuracy.
Another important question for transportation safety is whether the reported accident cause is consistent with narrative description. To address these questions, we applied deep learning methods together with powerful word embeddings such as Word2Vec and GloVe  to classify accident cause values for the primary cause field using the text in the narratives. The results show that such approaches can both accurately classify accident causes based on report narratives and find important inconsistencies in accident reporting\footnote{Source code is shared as an open source tool at \url{https://github.com/mojtaba-Hsafa/train_accidents}}.\\
\end{abstract}

\begin{IEEEkeywords} 
Rail accidents, Safety engineering, Text mining, Machine learning,  Neural networks
\end{IEEEkeywords} 

\section{Introduction}

Rail accident reporting in the U.S. has remained relatively unchanged for more than 40 years. The report form has 52 relevant accident fields and many of these fields have sub-fields. Some fields require entry of the value of an accident result or condition, e.g., ``Casualties to train passengers'' and ``Speed''. Other fields have restricted entries to values from a designated set of choices, e.g., ``Type of equipment'' and ``Weather''. ``Primary cause'' is an example of a restricted entry field in the report where the value must be one of 389 coded values. Choosing one of these categories while filling in reports is sometimes challenging and subject to errors due to the wide range of accidents. On the other hand, this field has significant importance for transportation administrations analysis in order to provide better safety regulations.  

Field 52 on the report is different from the other fields because it allows the reporter to enter a narrative description of the accident. These accident narratives provide a way for the accident reporter to describe the circumstances and outcomes of the accident in their own words. Among other things, this means providing details not entered in any of the other fields of the report. For example, while the report may show an accident cause of H401 - ``Failure to stop a train in the clear,'' the narrative could provide the reasons and circumstances for this failure. These additional details can be important in improving rail safety by helping in selection of a more accurate cause for the event. As a result, a method that correlates the detailed narratives with causes would be beneficial for both accident reporters and railroad administrators.

Despite the advantages of the narrative field, most safety changes result from fixed field entries since accident descriptions are difficult to automatically process. The advance of methods in text mining and machine learning has given us new capabilities to process and automatically classify textual content. This paper describes the results of a study using the latest methods in deep learning to process accident narratives, classify accident cause, and compare this classification to the causal entry on the form. The results of this study give us insights into the use of machine learning and, more specifically, deep learning to process accident narratives and to find inconsistencies in accident reporting.  

This paper investigates how the narrative fields of FRA accident reports could be efficiently used to extract the cause of accidents and establish a relationship between the narrative and the possible cause. Such relationships could assist the reporters to freely enter the narratives and getting candidate choices for causal field of reports. Our approach uses state-of-the-art deep learning technologies to classify texts based on their causes. The rest of this paper is organized as follows: in Section~\ref{Sec2}, related work in both accident analysis and text classification with deep learning have been presented. Section~\ref{Sec3} describes in detail the approach that has been used along with evaluation criteria. Section~\ref{Sec4} provides details of our implementations and section~\ref{Sec5} reports the results. Finally, Section~\ref{Sec6} presents the conclusion.

\section{Related Work}\label{Sec2}
This paper utilizes text mining and a new generation of natural language processing techniques, i.e. deep learning~\cite{kowsari2017HDLTex,kowsari2018rmdl,Heidarysafa2018RMDL} in an effort to discover relationships between accident reports' narratives and their causes. In this section, we describe related work in both railroad accident analysis and text mining with deep leaning.  
Train accident reports have been the subject of considerable research and different approaches have been used to derive meaningful information from these reports to help improve safety. As an example, the relationship between the length of train and accident rate has been investigated in~\cite{schafer2008relationship}. This paper also emphasizes the importance of proper causal understanding. Other authors~\cite{liu2011analysis,liu2012analysis} have used FRA data to investigate accidents caused by derailments. Recent work has used statistical analysis on FRA data to discover other patterns to investigate freight train derailment rate as an important factor \cite{liu2015statistical}. All  of these previous works used only the fixed field entries in the accident reports for their analysis and did not use information in the accident narratives. 
Some investigators have begun to apply text mining for accident report analysis in an attempt to improve safety. Nayak, et al.~\cite{nayak2010application} provided such an approach on crash report data between 2004 and 2005 in Queensland Australia. They used the Leximancer text mining tool to produce cluster maps and most frequent terms and clusters. Other research~\cite{jin2007improving}  introduced concept of chain queries that utilize text retrieval methods in combination with  link-analysis techniques. Recent work by Brown~\cite{brown2016text} provided a text analysis of narratives in accident reports to the FRA.  He specifically used topic modeling of narratives to characterize contributors to the accidents. 
In this paper, we present a new approach to the analysis of these accident narratives using deep learning techniques. We specifically applied three main deep learning architectures, Convolutional Neural Nets (CNN), Recurrent Neural Nets~(RNN), and Deep Neural Nets (DNN), to discover accident causes from the narrative field in FRA reports.

Another study~\cite{lecun2015deep} presented an overview of how these methods improved the state-of-the-art machine learning results in many fields such as object detection, speech recognition, drug discovery and many other applications. CNN were first introduced as a solution for problems involving data with multiple array structure such as 2D images. However, the researchers in~\cite{johnson2014effective}  proposed using a 1D structure to enable CNN applications for text classification. This work was extended by Zhang, et al., who developed character-level CNN for text classification~\cite{zhang2015character}. Other work has provided additional extensions to include use of dynamic  k-max pooling for the architecture in modeling sentences~\cite{blunsom2014convolutional}.
In RNN, the output from a layer of nodes can reenter as input to that layer. This architecture makes these deep learning models particularly suited for applications with  sequential data including, text mining. Irsoy et al.~\cite{irsoy2014opinion} showed an implementation of deep RNN structure for sentiment analysis of sentences. The authors of this paper compared their approach to the state-of-the-art conditional random fields baselines and showed that their method outperforms such techniques. Other researchers used different combinations of RNN models with some modifications and showed  better performance in document classifications as in~\cite{tang2015document},~\cite{lai2015recurrent}. Also, some recent researchers combined CNN and RNN in a hierarchical fashion and showed their overall improved performance for text classification as in~\cite{yang2016hierarchical}. Another hierarchical model for text classification is presented in~\cite{kowsari2017HDLTex}  where they employ stacks of deep learning architectures to provide improved document classification at each level of the document hierarchy. In our study, we have combined text mining methods and deep learning techniques to investigate the relationship of narrative field with accident cause which has not been explored before using such methods.
\begin{figure*}[t]
    \centering
    \includegraphics[width=\textwidth]{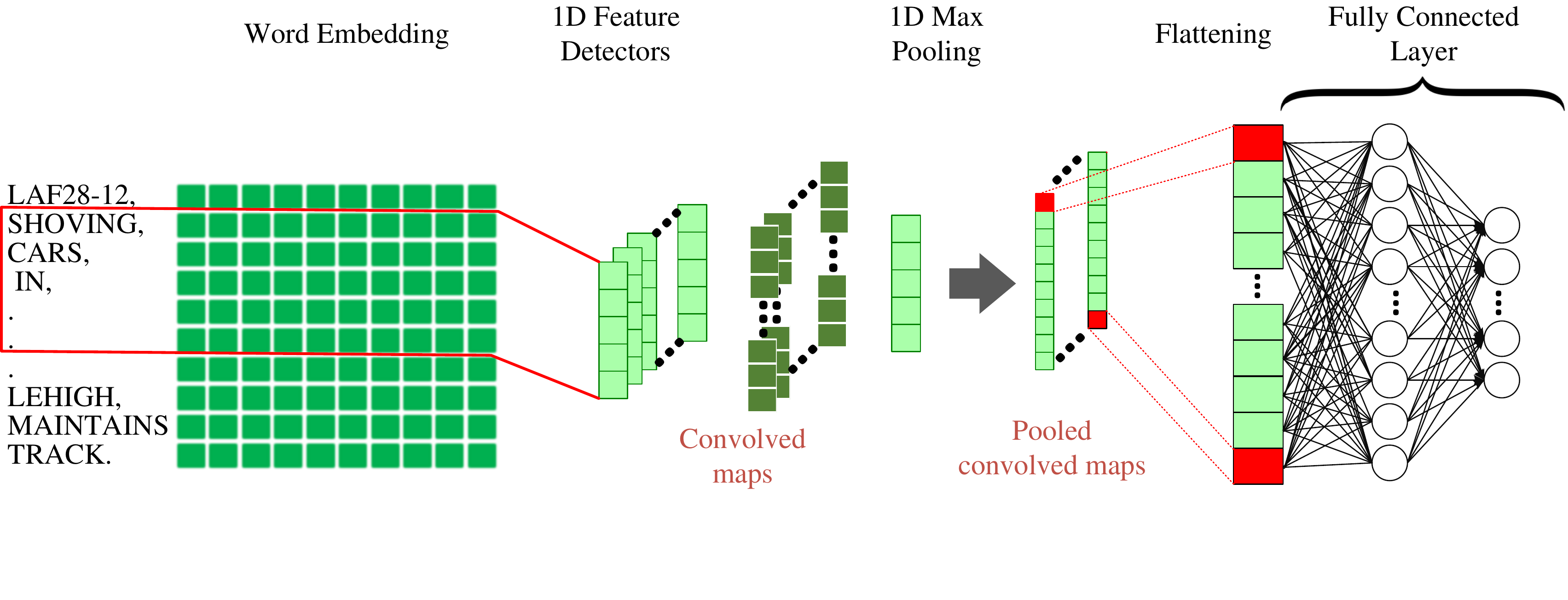}
    \vspace{-30pt}
    \caption{Structure of Convolutional Neural Net using  multiple 1D feature detectors and 1D max pooling} \label{cnn_fig}
\end{figure*}
\section{Method}\label{Sec3}
For this analysis, each report is considered as a single short document which consists of a sequence of words or unigrams. These sequences are considered input in our models and the accident cause (general category or specific coded cause) is the target for the deep learning model. We convert the word sequences into vector sequences to provide input to the deep learning models. Different solutions such as``Word Embedding'' and tf-idf representation are available to accomplish this goal. This section also provides details on deep learning architectures and evaluation methods used in this study .
\subsection{Word Embedding and Representation}
Different word representations have been proposed to translate words or unigrams into understandable numeric input for machine learning algorithms. One of the basic methods is term-frequency~(TF) where each word is mapped on to a number corresponding to the number of occurrences of that word in the whole corpora. Other term frequency functions present word frequency as a Boolean or a logarithmically scaled number. As a result, each document is translated to a vector containing the frequency of the words in that document. Therefore, this vector will be of the same length as the document itself. Although such an approach is intuitive, it suffers from the fact that common words tend to dominate the representation.
\subsubsection{Term Frequency-Inverse Document Frequency}
\textit{K. Sparck Jones}~\cite{sparck1972statistical} proposed  inverse document frequency (IDF) that can be used in conjunction with term frequency to lessen the effect of common words in the corpus. Therefore, a higher weight will be assigned to the words with both high frequency in a document and low frequency in the whole corpus. The mathematical representation of weight of a term in a document by tf-idf is given in~\ref{tf-idf} .
\begin{equation}\label{tf-idf}
    W(d,t)=TF(d,t)* log(\frac{N}{df(t)})
\end{equation}
Where~$N$ is the number of documents and~$df(t)$ is the number of documents containing the term $t$ in the corpus. The first part in Equation~\ref{tf-idf} would improve recall and the latter would improve the precision of the word embedding~\cite{tokunaga1994text}. Although tf-idf tries to overcome the problem of common terms in a document, it still suffers from some other descriptive limitations. Namely, tf-idf cannot account for similarity between words in the document since each word is presented as an index. In recent years, with development of more complex models such as neural networks, new methods have been presented that can incorporate concepts such as similarity of words and part of speech tagging. GloVe is one such word embedding technique that has been used in this work. Another successful word embedding method used in this work is Word2Vec which is described in the next part.

\subsubsection{Word2Vec} Mikolov, et al.  developed the \textit{``word to vector''} representation as a better word embedding approach~\cite{mikolov2013efficient}. Word2vec uses two neural networks to create a high dimensional vector for each word: Continuous Bag of Words~(CBOW) and continuous skip-gram~(CSG). CBOW represents the word in context with previous words while CSG represents the word by proximity in the vector space. Overall the word2vec method provides a very powerful relationship discovery approach.
\subsubsection{Global Vectors for Word Representation~(GloVe)}
\begin{figure}[]
    \centering
    \includegraphics[width=0.75\columnwidth]{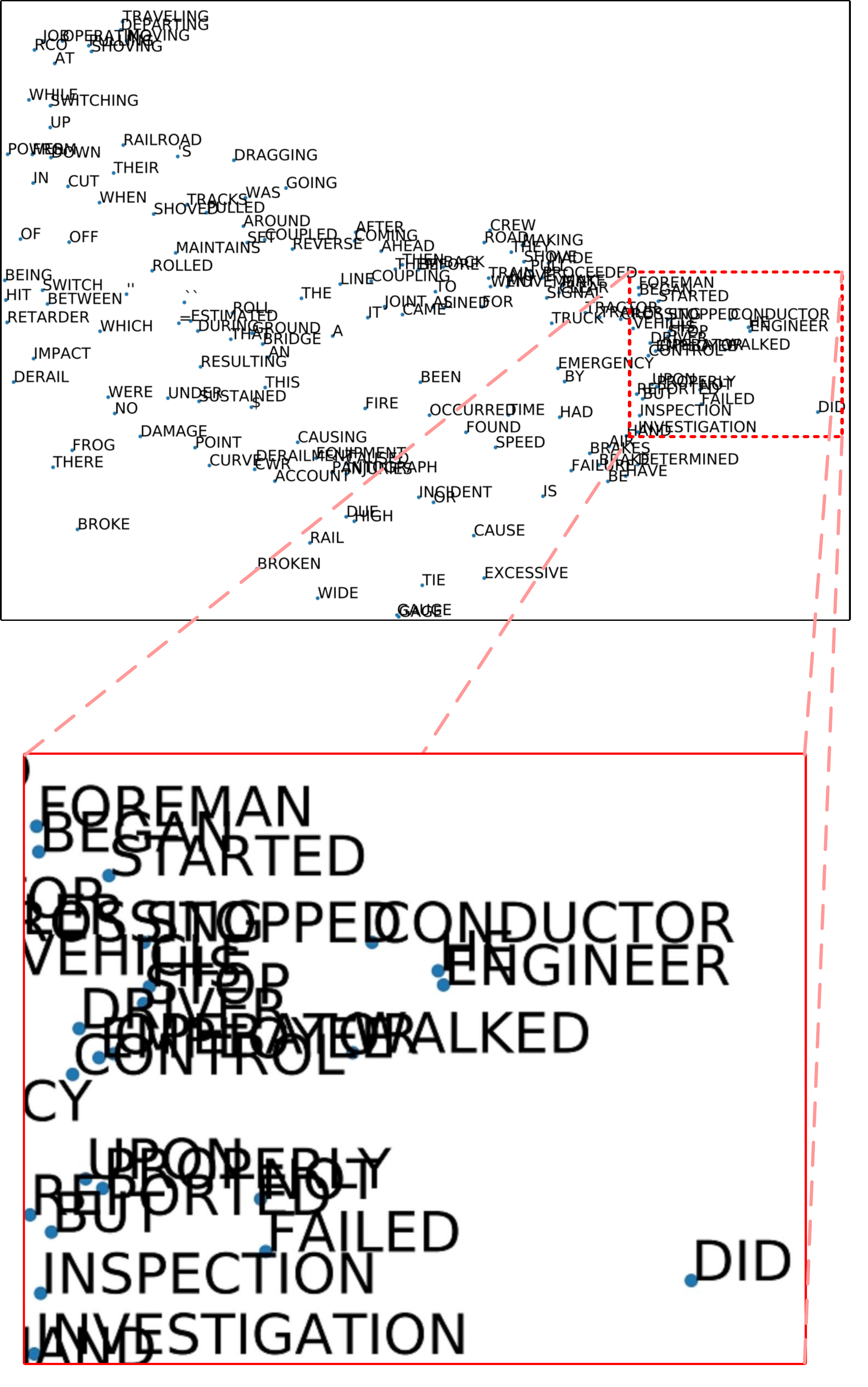}
    \caption{T-sne visualization of Word2vec 300 most common words }
    \label{fig:my_label}
\end{figure}

Another powerful word embedding technique is Global Vectors~(GloVe) presented in~\cite{pennington2014glove}. The approach is very similar to the word2vec method where each word is represented by a high dimension vector, and trained based on the surrounding words over a huge corpus. The pre-trained embeddings for words used in this work are based on 400,000 vocabularies trained over Wikipedia 2014 and Gigaword 5 with 50 dimensions for word representation. GloVe also provides other pre-trained word vectorizations with 100, 200, 300 dimensions which are trained over even bigger corpi as well as over Twitter. Figure \ref{fig:my_label} shows an example of how these embeddings can be used to transfer words to a better representation. As one can see, words such as ``Engineer'', ``Conductor'', and ``Foreman'' are considered close based on these embeddings. Similarly, words such as ``inspection'' and ``investigation'' are considered very similar.  

\subsection{Text Classification with Deep Learning}\label{deepClass}
Three deep learning architectures used in this paper to analyze accident narratives, are Convelutional Neural Networks (CNN), Recurrent Neural Networks (RNN), and Deep Neural Networks (DNN)~\cite{kowsari2017HDLTex,kowsari2018rmdl,Heidarysafa2018RMDL}. The building blocks of these classifiers are described in greater detail in this section.

\subsubsection{Deep Neural Networks (DNN)}\label{subsubsec:DNN}
DNN's structure is designed to learn by multiple connections between layers where each layer only receives connections from previous layer and provides connections only to the next layer ~\cite{kowsari2018rmdl,Heidarysafa2018RMDL}. The input is a vectorized representation of documents, which connects to the first layer. The output layer is number of classes for multi-class classification and only one output for binary classification.   
The implementation of Deep Neural Networks~(DNN) is discriminative trained model that uses standard back-propagation algorithm. Different activation functions for nodes exist such as sigmoid or tanh but we noticed ReLU~\cite{nair2010rectified}~(Equation~\ref{relu}) provides better results. The output layer for multi-class classification, should use $Softmax$ as shown in Equation~\ref{Softmax}.
\begin{align}
f(x) &= \max(0,x)\label{relu}\\
\sigma(z)_j &= \frac{e^{z_j}}{\sum_{k=1}^K e^{z_k}}\label{Softmax}\\ 
&\forall   ~j \in \{1,\hdots, K\} \nonumber
\end{align}

 Given a set of example pairs $(x,y),x\in X,y\in Y $, the goal is to learn from these input and target spaces using hidden layers. In our text classification, the input is a string which is generated by vectorization of text using tf-idf word weighting. 
 
\subsubsection{Convolutional Neural Nets}
Convolutional neural networks~(CNN) were introduced by Yann Lecun~\cite{lecun1998gradient} to recognize handwritten digits in images. The proposed design, though powerful, did not catch the attention of the computer-vision and machine learning communities until almost a decade later when higher computation technologies such as Graphics Processing Units (GPU) became available~\cite{lecun2015deep}.  Although CNNs have been designed with the intention of being used in the image processing domain, they have also been used in text classification using word embedding~\cite{kim2014convolutional,kowsari2018rmdl,Heidarysafa2018RMDL}. 

In CNN, a convolutional layer contains connections to only a subset of the input. These subsets of neurons are \textit{receptive fields} and the distance between receptive fields is called \textit{stride}. The value at any neuron in the receptive field is given by the output from an activation function applied to the weighted sum of all inputs to the receptive field. Common choices for activation functions are sigmoid, hyperbolic tangent, and rectified linear. As with most CNN architectures, in this study we stack multiple convolutional layers on top of each other.

The next structure in the CNN architecture is a pooling layer. The neurons in this layer again sample a small set of inputs to produce their output value. However, in this case they simply return the minimum, average or maximum of the input values. Pooling reduces computation complexity, and memory use. Additionally, it can improve performance on translated and rotated inputs~\cite{scherer2010evaluation}. Pooling can be repeated multiple times depending on the size of input and the complexity of the model. 

The final layer is traditional fully connected layers taking a flattened output from the last pooling layer as its input. The output from this fully connected network is run through a softmax function for multinomial (i.e., multiple labels) problems, such as classifying cause from accident narratives. 

Figure~\ref{cnn_fig} shows the structure of an example CNN with one convolutional and max pooling layer for text analysis.
\begin{figure}[t]
\centering
\includegraphics[width=0.95\columnwidth]{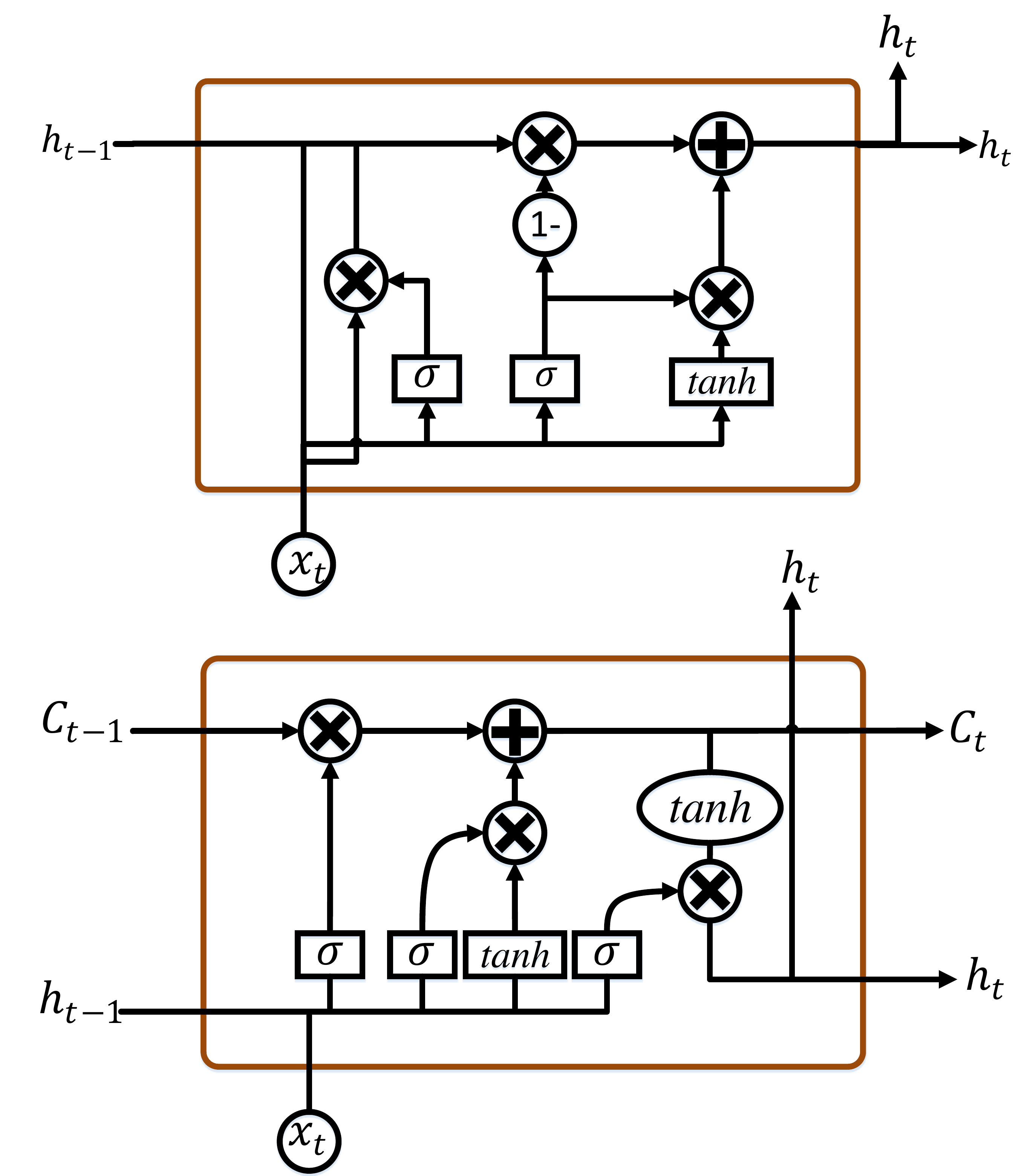}
\caption{Top Fig: A cell of GRU, Bottom Fig: A cell of LSTM~\cite{Heidarysafa2018RMDL}}\label{fig:LSTM}
\end{figure}
\subsubsection{Recurrent Neural Networks~(RNN)}
RNN are a more recent category of deep learning architectures where outputs are fed backward as inputs. Such a structure allows the model to keep a memory of the relationship between words in nodes. As such, it provides a good approach for text analysis by keeping sequences in memory~\cite{karpathy2015unreasonable}. 

The general RNN structure is formulated as in Equation~\ref{rnn_gen} where $x_t$ denotes the state at time $t$ and $\boldsymbol{u_t}$ refers to the input at step $t$.
\begin{equation}
\label{rnn_gen}
x_{t}=F(x_{t-1},\boldsymbol{u_t},\theta)
\end{equation}
Equation~\ref{rnn_gen} can be expanded using proper weights as shown in Equation~\ref{rnn_spec}.
\begin{equation}\label{rnn_spec}
x_{t}=\mathbf{W_{rec}}\sigma(x_{t-1})+\mathbf{W_{in}}\mathbf{u_t}+\mathbf{b}.
\end{equation}
In this equation $\mathbf{W_{rec}}$ is the recurrent matrix weight, $\mathbf{W_{in}}$ are the input weights, $\mathbf{b}$ is the bias, and $\sigma$ is an element-wise function. 

The general RNN architecture has problems with vanishing and, less frequently, exploding gradients. This happens when the gradient goes through the recursions and gets progressively smaller or larger in vanishing or exploding states respectively.~\cite{bengio1994learning}. To deal with these problems, long short-term memory~(LSTM), a special type of RNN that preserves long-term dependencies was introduced which shows to be particularly effective at mitigating the vanishing gradient problem~\cite{pascanu2013difficulty}.\\

Figure~\ref{fig:LSTM} shows the basic cell of an LSTM model. Although LSTM has a chain-like structure similar to RNN, LSTM uses multiple gates to regulate the amount of information allowed into each node state~\cite{kowsari2017HDLTex,kowsari2018rmdl,Heidarysafa2018RMDL}.
\begin{figure}[h]
    \centering
    \includegraphics[width=\columnwidth]{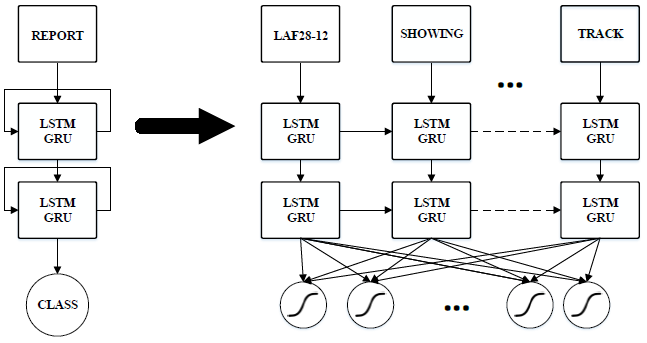}
    \caption{Structure of Recurrent Neural Net for report analysis using two LSTM/GRU layers}
    \label{fig:General_rnn}
\end{figure}

\textit{\textbf{Gated Recurrent Unit~(GRU)}}\label{subsec:GRU}
The Gated Recurrent Unit~(GRU)~\cite{cho2014learning} is a more recent and simpler gating mechanism than LTSM. GRU contains two gates, does not possess internal memory (the $C_{t-1}$ in Figure~\ref{fig:LSTM}), and unlike LSTM, a second non-linearity is not applied (tanh in Figure~\ref{fig:LSTM}). We used GRU as our main RNN building block. A more detailed explanation of a GRU cell is given in following:
\begin{equation}
z_{t}=\sigma_g(W_{z}x_{t}+U_zh_{t-1}]+b_{z}), \label{eq:gru1}
\end{equation}
Where~$z_t$ refers to update gate vector of~$t$,~$x_t$ stands for input vector,~$W$, $U$ and~$b$ are parameter matrices and vector, $\sigma_g$ is the activation function, which could be sigmoid or ReLU.
\begin{equation}
    \tilde{r_{t}}=\sigma_g(W_{r}x_{t}+U_rh_{t-1}]+b_{r}), \label{eq:gru2}
\end{equation}
Where~$r_t$ stands for the reset gate vector of~$t$.
\begin{equation}
   h_t =  z_t \circ h_{t-1} + (1-z_t) \circ \sigma_h(W_{h} x_t + U_{h} (r_t \circ h_{t-1}) + b_h)\label{eq:gru6}
\end{equation}
Where~$h_t$ is output vector of~$t$, $r_t$ stands for reset gate vector of~$t$, $z_t$ is update gate vector of~$t$, $\sigma_h$ indicates the hyperbolic tangent function.

Figure~\ref{fig:General_rnn} shows the RNN architectures used in this study by employing either LSTM or GRU nodes.

\subsection{Evaluation}
In order to understand how well our model performs, we need to use appropriate evaluation methods to overcome problems such as unbalanced classes. This section describes our evaluation approach.
\begin{table*}[]
\centering
\caption{Distribution of data point and specified categories according  to FRA}
\label{specfic_dstro}
\begin{tabular}{|c|c|c|c|c|c|c|c|c|}
\hline
 Total reports& 'H306-7'&'T110'& 'H702'&'T220-207'& 'T314'&  'M405'& 'H704'& 'H503'\\ \hline
11982&2613& 2448& 2171& 1716& 1053& 753 & 652 & 576\\ \hline

\end{tabular}
\end{table*}

\subsubsection{F1 measurement}
With unbalanced classes, as with accident reports, simply reporting the overall accuracy would not reflect the reality of a model's performance. For instance, because some of these classes have considerably more observations than others, a classifier that chooses these labels over all others will obtain high accuracy, while misclassifying the smaller classes. Hence, the analysis in this paper requires a more comprehensive metric. One such metric is F1- score and its two main implementations:   Macro-averaging and Micro-averaging.  The macro averaging formulation is given in Equations~\ref{F1macro2},  using the definition of precision ($\pi$) and recall ($\rho$) in Equation \ref{pres},\ref{recall}.
\begin{align}
    \pi_{i}&=\frac{TP_{i}}{TP_{i}+FP_{i}}\label{pres}\\
    \rho_{i}&=\frac{TP_{i}}{TP_{i}+FN_{i}}\label{recall}\\
    F_{i}&= \frac{2\pi_{i}\rho_{i}}{\pi_{i}+\rho_{i}}\\
    F_{1-macro}&= \frac{\sum_{i=1}^{N}F_{i}}{N}\label{F1macro2}
\end{align}
Here $TP_{i}$, $FP_{i}$, $TN_{i}$ represent true positive, false positive and true negative, respectively, for class $i$ and $N$ classes.

Our analysis uses macro averaging which tends to be biased toward less populated classes~\cite{ozgur2005text}. As a result, we provide a more conservative evaluation since deep learning methods tend to perform worse with smaller data sets.
Another performance measure used in this study, is confusion matrix. A confusion matrix compares true values with predicted values and therefore, provides information on which classes are mostly misclassified to what other classes .
\section{Experiments}\label{Sec4}
In this section, we describe the embeddings that are used for our analysis as well as the structure of each deep learning model and the hardware that has been used to perform this work. 
To create word2vec presentation, we used gensim library to construct a 100 dimension vector for each word using a window of size 5. Similarly, we used a 100 dimension representation of Glove trained over 400K vocabulary corpus. The input documents have been padded to be of the same size of 500 words for all narratives. Our experiments showed that higher dimensions would not have a significant effect on the results.

Our DNN implementation consists of five hidden layers, where in each hidden layer, we have~$1000$ units with ReLu activation function followed by a dropout layer.

Our CNN implementation consists of three 1D convolutional layers, each of them followed by both a max pool and dropout layer. Kernel size for convolution and max pooling layers was both 5. At the final layer our fully connected layer has been made from 32 nodes and used a dropout layer as well.

RNN implementation is made of two GRU layers with 64 units in each followed by dropout after them. Final layer is a fully connected layer with 64-128 nodes at the end. This layer also includes a dropout similar to previous layers. The dropout rate is between 0.1 to 0.5 depending on the task and model which helps to reduce the chance of overfitting for our models.\\
The processing was done on a $Xeon~E5-2640~ (2.6 GHz)$ with $32$ cores and $64 GB$ memory. We used Keras package~\cite{chollet2015keras} with Tensorflow as its backend for our implementation.

\begin{figure}[t]
  \centering
  \includegraphics[width=\linewidth]{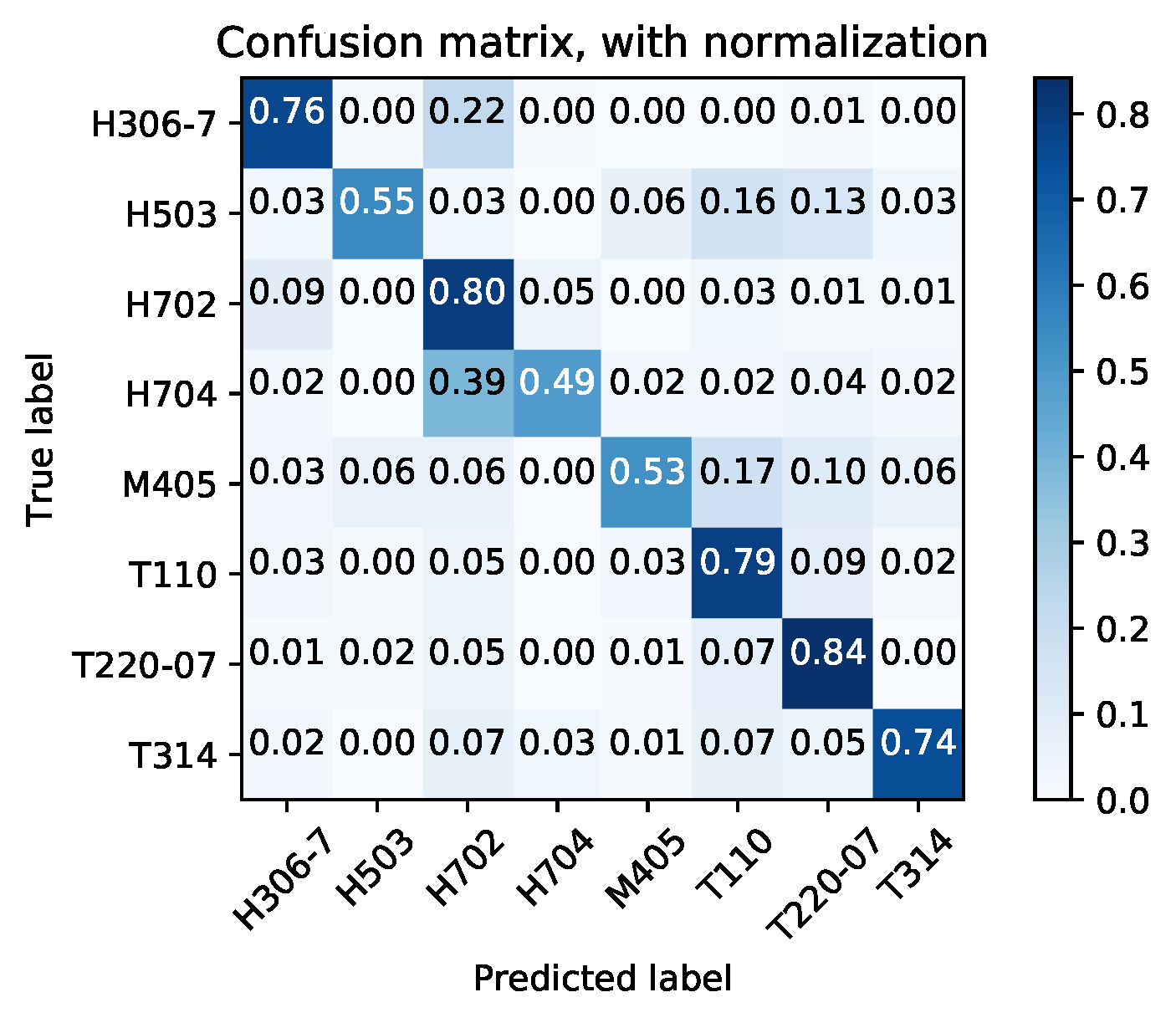}
  \label{fig:sub2}
\caption{Confusion matrix for the best classifier}
\label{fig:specific matrix}
\end{figure}

\section{Results}\label{Sec5}
This work has been performed using  Federal Railroad Administration (FRA) reports collected during 17 consecutive years (2001-2017)~\cite{FRAonline}. 
FRA provides a narrative for each accident with the corresponding cause reported on that accident.
The results are in two sections. In the first section, we show the performance in labeling the general cause for each accident based on its narrative and in the second section, we focus on the specific accident cause, on most common type of accidents according to reported detailed cause. In both of these analyses, we also compare our performance with some of traditional machine learning algorithms such as Support Vector Machines (SVM), Naive Bayes Classifier (NBC) and Random Forest as our baselines. Finally, we look at our misclassified results using confusion matrix and analyze errors made by our models.

\begin{figure}[t]
\centering
\includegraphics[width=0.85\linewidth]{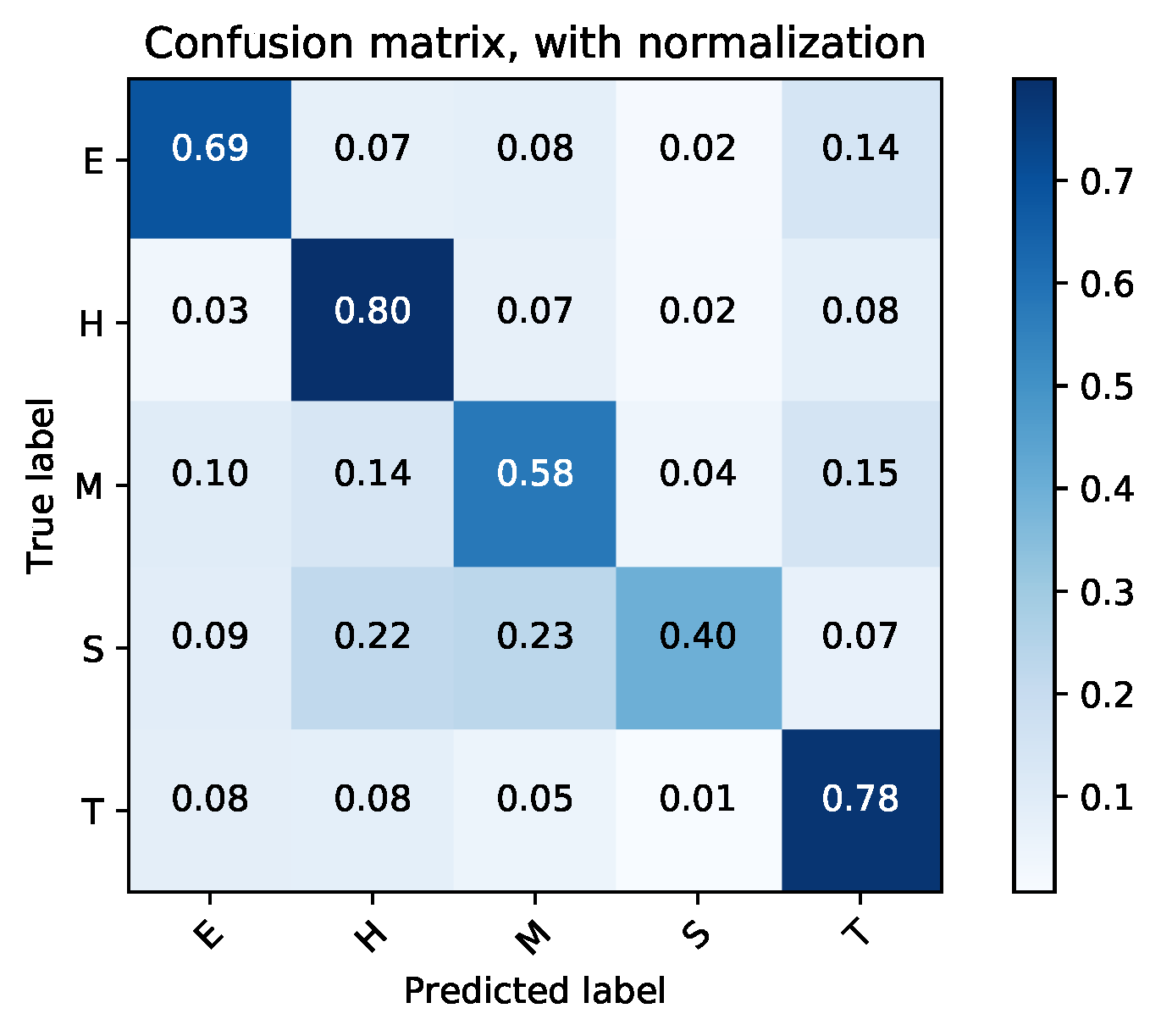}
  \label{fig:sub3}
\caption{Confusion matrix for the best classifier}
\label{fig:general matrix}
\end{figure}

\subsection{General cause analysis}
The general accident cause is in the reported cause field of accident reports. This analysis considers
$40,164$ reports with five labels as general causes.  Table~\ref{ta_data} shows the five causal labels and their distribution.
 \begin{table}
\centering
\caption{Distribution of data points and general labels~(E: Electrical failure,  H: Human Factor,  M: Miscellaneous, S: Signal communication, and  T: Track)}
\label{ta_data}
\begin{tabular}{|c|c|c|c|c|c|}
\hline
 Total reports& E  &  H   &  M   & S  &  T\\ \hline
40164 & 5118      & 15152    & 5762   & 786        & 13256        \\ \hline
\end{tabular}
\end{table}

To classify the reports, both RNN and CNN along with two word embeddings, Word2Vec and Glove, and DNN with tf-idf are used. Table~\ref{result_general} shows the performance of our techniques and compare it with our baselines. Generally, Word2Vec embedding produces better F1 scores over the test set. Also, the differences between RNN and CNN results are not significant. 

Figure~\ref{fig:general matrix} shows the confusion matrix for the best classifier. This confusion matrix shows that deep learning models in conjunction with vector representations of words can provide good accuracy especially on categories with more data points. 

\begin{figure}[t]
\centering
\begin{minipage}{\linewidth}
  \centering
  \includegraphics[width=\linewidth]{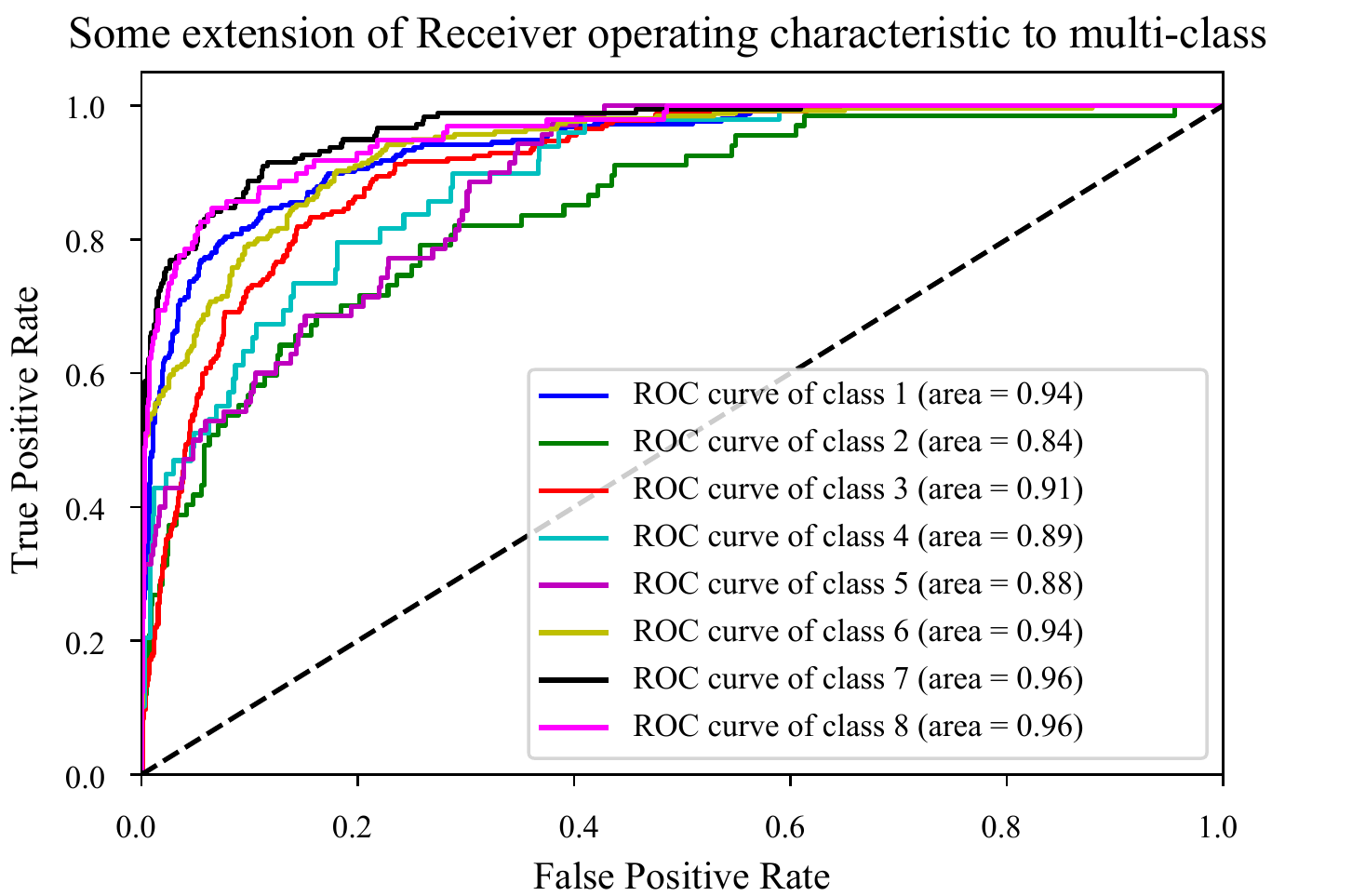}
  \label{fig:roc_d}
\end{minipage}%
\\
\begin{minipage}{\linewidth}
  \centering
  \includegraphics[width=\linewidth]{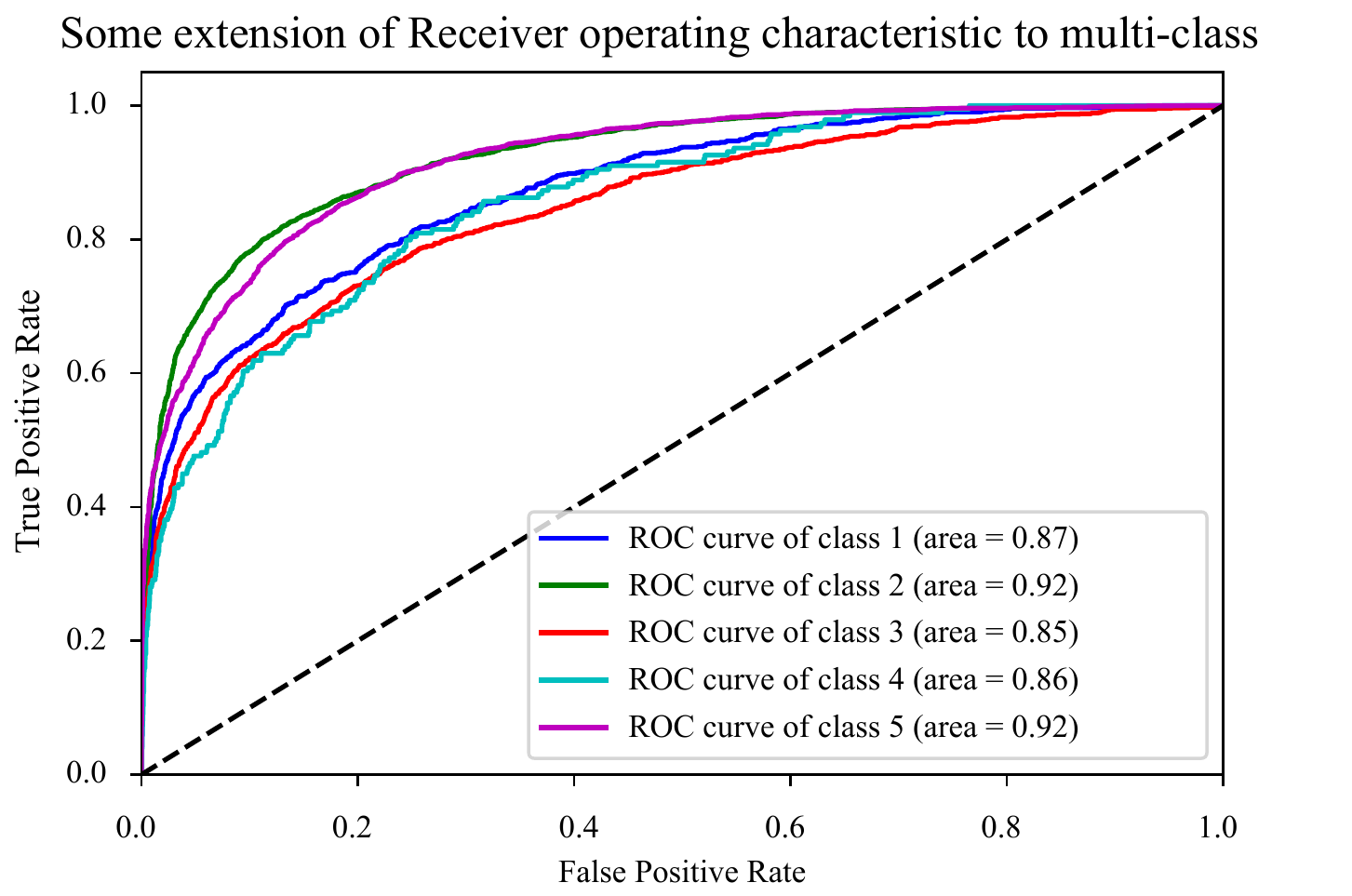}
  \label{fig:roc_}
\end{minipage}
\caption{ROC curves for classifier of general and specific causes}
\label{fig:ROC}
\end{figure}

\begin{table}[]
\centering
\caption{Classification F1 score of combined techniques}
\label{result_general}
\begin{tabular}{|c|c|c|c|}
\hline
\multicolumn{2}{|c|}{Feature Extraction}   & Technique & \multicolumn{1}{c|}{\begin{tabular}[c]{@{}c@{}}F1 measure\\ (Macro)\end{tabular}} \\ \hline
\multirow{4}{*}{Word Embedding} & Word2Vec    & CNN       & 0.65                  \\ \cline{2-4} 
                                & Word2Vec    & RNN       & 0.64                  \\ \cline{2-4} 
                                & GloVe     & CNN       & 0.63                 \\ \cline{2-4} 
                                & Glove     & RNN       & 0.59                 \\ \hline
\multirow{4}{*}{Word Weighting} & tf-idf   & DNN       & 0.61               \\ \cline{2-4} 
                                & tf-idf   & SVM       & 0.57                \\ \cline{2-4} 
                                & tf-idf   & NBC       & 0.61                 \\ \cline{2-4} 
                                & tf-idf  & Random Forest        & 0.57                 \\ \hline
\end{tabular}
\end{table}

\subsection{Specific cause analysis}
Our analysis also considers more specific accident causes in FRA reports (one of 389 code categories).  An obvious issue with more detailed causal labels is that there are some cause categories with very few reports. Therefore, over the period studied, the top ten most common causes (combined into 8 categories since H307 and H306 have the same description and the description of T220 and T207 is very similar) have been selected for analysis. Table~\ref{specfic_dstro} shows the distribution of reports on these categories. Figure~\ref{fig:specific matrix} shows the confusion matrix for the best classifier for the top 8 categories of causes.

We also investigate classifier performance using ROC curves as in Figure~\ref{fig:ROC} for both general and specific causes.

Table~\ref{result_specific} shows the results for specific causes along with a comparison with our baselines' performances. Similar to our previous results, models using Word2Vec embedding perform better than the ones using GloVe both in CNN and RNN architecture. 

\subsection{Error analysis}
To better understand model performance, we investigated the errors made by our classifiers. The confusion matrices, clearly show that the number of instances in the classes plays a major role in classification performance. 

As an example, reports labeled with Signal as the main cause are the smallest group and not surprisingly, the model does poorly on these reports due to the small number of training data points. 

There is, however, another factor at work in model performance which comes from rare cases where the description seems uncorrelated to the cause. As an example of such cases, our model predicted the following narrative "DURING HUMPING OPERATIONS THE HOKX112078 DERAILED IN THE MASTER DUE TO EXCESSIVE RETARDER FORCES." in mechanical category while the original category reported is cause by Signal. This seems not consistent with the report narrative.

Identifying such inconsistencies in reports' narratives is important because both policy changes and changes to operation result from aggregate analysis of accident reports. 

 \begin{table}[]
\centering
\caption{Classification F1 score of combined techniques for specific causes}
\label{result_specific}
\begin{tabular}{|c|c|c|c|}
\hline
\multicolumn{2}{|c|}{Feature Extraction}   & Technique & \multicolumn{1}{c|}{\begin{tabular}[c]{@{}c@{}}F1 measure\\ (Macro)\end{tabular}}  \\ \hline
\multirow{4}{*}{Word Embedding} & Word2Vec    & RNN       & 0.71                  \\ \cline{2-4} 
                                & Word2Vec    & CNN       & 0.66                 \\ \cline{2-4} 
                                & GloVe & RNN       & 0.64                  \\ \cline{2-4} 
                                & GloVe & CNN       & 0.62                  \\ \hline
\multirow{4}{*}{Word Weighting} & tf-idf   & DNN       & 0.64                 \\ \cline{2-4} 
                                & tf-idf   & SVM       & 0.61                \\ \cline{2-4} 
                                & tf-idf   & NBC       & 0.33                 \\ \cline{2-4} 
                                & tf-idf  & Random Forest        & 0.62                 \\ \hline
\end{tabular}
\end{table}

\section{Conclusion and Future Work}
\label{Sec6}
This paper presents deep learning methods that use the narrative fields of FRA reports to discover the cause of each accident. These textual fields are written using specific terminologies, which makes the interpretation of the event cumbersome for non-expert readers. However, our analysis shows that when using proper deep learning models and word embeddings such as GloVe and especially Word2Vec, the relationship between these texts and the cause of the accident could be extracted with acceptable accuracy. The results of testing for the five major accident categories and top 10 specific causes (according to FRA database coding) show the deep learning methods we applied were able to correctly classify the cause of a reported accident with overall 75
\% accuracy. Also, the results indicate that applying recent deep learning methods for text analysis can help exploit accident narratives for information useful to safety engineers. This can be done by providing an automated assistant that could help identify the most probable cause of an accident based on the event narrative.
Also, these results suggest that in some rare cases, narrative description seems inconsistent with the suggested cause in the report. Hence, these methods may have promise for identifying inconsistencies in the accident reporting and thus could potentially impact  safety regulations.  Moreover, the classification accuracy is higher in more frequent accident categories. This suggests that as the number of reports increases, the accuracy of deep learning models improves and these models become  more helpful in interpreting such domain specific texts.




\section*{Acknowledgement}
This work was supported by The United States Army Research Laboratory under Grant W911NF-17-2-0110.

\bibliographystyle{IEEEtran}
\bibliography{ref}

\end{document}